\documentclass[11pt,a4paper]{article}
\usepackage[nohyperref]{acl2018} % erroreak ematen ditu!!
\usepackage{times}
\usepackage{latexsym}

\usepackage{url}

\aclfinalcopy % Uncomment this line for the final submission
\usepackage[disable]{todonotes} % notes not showed

\usepackage[utf8]{inputenc}
\usepackage{multirow}
\usepackage{url}
\usepackage{booktabs}			% Pretty tables
\usepackage{color}
\usepackage{fancyvrb}

\title{The risk of sub-optimal use of Open Source NLP Software: \\ 
UKB is inadvertently state-of-the-art in knowledge-based WSD}

\author{Eneko Agirre \\
  IXA NLP group \\
  UPV/EHU \\ \And
  Oier López de Lacalle \\
  IXA NLP group \\
  UPV/EHU \\
  {\tt \{e.agirre,oier.lopezdelacalle,\textbf{a.soroa}\}@ehu.eus} \\\And
  Aitor Soroa \\
  IXA NLP group \\
  UPV/EHU 
}

\date{}

\begin{document}
\maketitle
\begin{abstract}

  UKB is an open source collection of programs for performing, among
  other tasks, knowledge-based Word Sense Disambiguation (WSD). Since
  it was released in 2009 it has been often used out-of-the-box in
  sub-optimal settings. We show that nine years later it is the
  state-of-the-art on knowledge-based WSD. This case shows the
  pitfalls of releasing open source NLP software without optimal
  default settings and precise instructions for reproducibility.
  % This paper provides the details for optimal
  % performance in WSD, updates results in some later datasets and
  % compares to the latest state-of-the-art, where it performs best
  % among knowledge-based systems.  The paper accompanies additional
  % software for full reproducibility.
\end{abstract}

\section{Introduction}
\label{sec:introduction}

The release of open-source Natural Language Processing (NLP) software
has been key to make the field progress, as it facilitates other
researchers to build upon previous results and software easily. It
also allows easier reproducibility, allowing for sound scientific
progress. Unfortunately, in some cases, it can also allow competing
systems to run the open-source software out-of-the-box with
sub-optimal parameters, specially in fields where there is no standard
benchmark and new benchmarks (or new versions of older benchmarks) are created.

Once a paper reports sub-optimal results for a NLP software, newer
papers can start to routinely quote the low results from the previous
study. Finding a fix to this situation is not easy. The authors of the
software can contact the authors of the more recent papers, but it is
usually too late for updating the paper. Alternatively, the authors of
the NLP software can try to publish a new paper with updated results,
but there is usually no venue for such a paper, and, even if
published, it might not be noticed in the field.

In this paper we want to report such a case in Word Sense Disambiguation (WSD), where the original
software (UKB) was released with suboptimal default parameters. Although the
accompanying papers did contain the necessary
information to obtain state-of-the-art results, the software did not contain step-by-step instructions,
or end-to-end scripts for optimal performance. This case is special,
in that we realized that the software is able to attain
state-of-the-art results also in newer datasets, using the same settings as in the papers.

The take-away message for open-source NLP software authors is that
they should not rely on other researchers reading the papers with
care, and that it is extremely important to include, with the software
release, precise instructions and optimal default parameters, or
better still, end-to-end scripts that download all resources, perform
any necessary pre-processing and reproduce the results. 

The first section presents UKB and WSD, followed by the settings and
parameters. Next we present the results and comparison to the
state-of-the-art. Section \ref{sec:results} reports some additional
results, and finally, we draw the conclusions. 

\section{WSD and UKB}
\label{sec:wsd-ukb}

Word Sense Disambiguation (WSD) is the problem of assigning the
correct sense of a word in a context
\citep{Agirre:2007:WSD:1564561}. Traditionally, supervised approaches
have attained the best results in the area, but they are expensive to
build because of the need of large amounts of manually annotated
examples. Alternatively, knowledge based approaches rely on lexical
resources such as WordNet, which are nowadays widely available
in many languages~\citep{bond2012survey}\footnote{\url{http://compling.hss.ntu.edu.sg/omw/}}. In
particular, graph-based approaches represent the knowledge base as a
graph, and apply several well-known graph analysis algorithms to
perform WSD.

UKB is a collection of programs which was first released for
performing graph-based Word Sense Disambiguation using a pre-existing
knowledge base such as WordNet, and attained state-of-the-art results
among knowledge-based systems when evaluated on standard
benchmarks~\citep{Agirre:09a,DBLP:journals/coling/AgirreLS14}. %qtleap?
In addition, UKB has been extended to perform
disambiguation of medical entities~\citep{ASS2010}, named-entities~\citep{tac12,arxiv}, word similarity~\citep{agirre2009naacl}
and to create knowledge-based word embeddings~\citep{goikoetxea2015}. All programs are open source\footnote{\url{http://ixa2.si.ehu.eus/ukb}}\textsuperscript{,}\footnote{\url{https://github.com/asoroa/ukb}} and are  accompanied by
the resources and instructions necessary to reproduce the results. The software is quite popular, with 60 stars and 26 forks in github, as well as more than eight thousand direct downloads from the website since 2011. The software is coded in C++ and released under the GPL v3.0 license.

%\item Downloads: 10.000 baina gehiago (estatistikak 2011tik ditugu soilik,
%  eta 8.200 inguru download daude). Orria hemen dago:
%
%  http://ixa2.si.ehu.es/awstats/awstats.pl

When UKB was released, the papers specified the optimal parameters for WSD \citep{Agirre:09a,DBLP:journals/coling/AgirreLS14}, as well as other key issues like the underlying knowledge-base version, specific set of relations to be used, and method to pre-process the input text. At the time, we assumed that future researchers would use the optimal parameters and settings specified in the papers, and that they would contact the authors if in doubt. The default parameters of the software were not optimal, and the other issues were left under the users responsibility. 

The assumption failed, and several papers 
reported low results in some new datasets (including
updated versions of older datasets), as we will see in the following
sections.

%\todo{inline}In this paper, we analyze the best set of parameters for obtaining an
%optimal performance using UKB. Along with this paper, we release UKB version
%3.1, which contains scripts to automatically reproduce the results of this
%paper.

%\label{sec:pre-processing}

\begin{table*}[t]
  \centering
  \begin{tabular}{lr|rrrrr}
                 & All           & S2 & S3 &
S07 & S13 & S15  \\
    \hline
      UKB (this work)    & \textbf{67.3} & 68.8          & 66.1          & 53.0          & \textbf{68.8} & \textbf{70.3} \\
      UKB (elsewhere)$\dagger\ddagger$ & 57.5          & 60.6          & 54.1          & 42.0          & 59.0 & 61.2 \\ \hline
      \citet{aaai18} $\ddagger$ & 66.9          & \textbf{69.0} & \textbf{66.9} & 55.6          & 65.3          & 69.6          \\
      Babelfy   \cite{Moro:2014:ELmeetsWSD}$\dagger$   & 65.5          & 67.0          & 63.5          & 51.6          & 66.4          & 70.3          \\
      MFS          & 65.2          & 66.8          & 66.2          & 55.2          & 63.0          & 67.8          \\
      \citet{conf/coling/BasileCS14}$\dagger$  & 63.7          & 63.0          & 63.7          & \textbf{56.7} & 66.2          & 64.6          \\
      \citet{Banerjee:2003:EGO:1630659.1630775}$\dagger$  & 48.7          & 50.6          & 44.5          & 32.0          & 53.6          & 51.0          \\
      \hline
\end{tabular}
  \caption{F1 results for knowledge-based systems on the~\cite{E17-1010} dataset. Top rows show conflicting results for UKB. $\dagger$ for results reported in ~\cite{E17-1010}, $\ddagger$ for results reported in \cite{aaai18}. Best results in bold. S2 stands for Senseval-2, S3 for Senseval-3, S07 for Semeval-2007, S13 for Semeval-2013 and S15 for Semeval-2015. }
  \label{tab:comparison}
\end{table*}

\begin{table*}[t]
  \centering
  \begin{tabular}{lr|rrrrr}
                 & All           & S2 & S3 &
S07 & S13 & S15  \\
    \hline
      \citet{C16-1130}   & \textbf{71.5}  & \textbf{73.8}   & \textbf{71.8}  & {63.5}  & \textbf{69.5}  & \textbf{72.6}          \\
      \citet{raganato-dellibovi-navigli:2017:EMNLP2017} & 69.9 & 72.0 & 69.1 & \textbf{64.8}& 66.9&71.5 \\ 
      \citet{iacobacci-pilehvar-navigli:2016:P16-1}$\dagger$  & 69.7          & 73.3          & 69.6          & 61.1          & 66.7          & 70.4 \\
      \citet{Melamud2016context2vecLG}$\dagger$    & 69.4          & 72.3          & 68.2          & 61.5          & 67.2          & 71.7          \\
      IMS \cite{Zhong:2010}$\dagger$    & 68.8          & 72.8          & 69.2          & 60.0          & 65.0          & 69.3          \\  \hline
%      IMS \cite{Zhong:2010}$\dagger$    & 68.3          & 70.8          & 68.9          & 58.5          & 66.3          & 69.7          \\  \hline
\end{tabular}
  \caption{F1 results for supervised systems on the~\cite{E17-1010} dataset. $\dagger$ for results reported in ~\cite{E17-1010}. Best results in bold. Note that \cite{raganato-dellibovi-navigli:2017:EMNLP2017} used S07 for development. }
  \label{tab:comparison2}
\end{table*}

\section{UKB parameters and setting for WSD}
\label{sec:ukb-param-sett}

When using UKB for WSD, %it can be parametrized by a large set of
%parameters and options, which greatly affect the final outcome of the application. 
the main parameters and settings can be classified in five main
categories. For each of those we mention the best options and the associated UKB parameter when relevant (in italics), as taken from 
\cite{Agirre:09a,DBLP:journals/coling/AgirreLS14}:
\begin{itemize}
\item Pre-processing of input text. When running UKB for WSD, one
  needs to define which window of words is to be used as context to
  initialize the random walks. One option is to take just the
  sentence, but given that in some cases the sentences are very short,
  better results are obtained when considering previous and following
  sentences. The procedure in the original paper repeated the extension procedure until
  the total length of the context is at least $20$ words\footnote{The number
    $20$ was initial arbitrarily set in the experiments
    of~\cite{Agirre:09a} somewhat arbitrarily, and never changed
    afterwards.}.
\item Knowledge base relations. When performing WSD for English, UKB
  uses WordNet~\cite{fellbaum98wordnet} as a knowledge base. WordNet
  comes in various versions, and usually UKB performs best when using
  the same version the dataset was annotated with. Besides regular WordNet
  relations, gloss relations (relations between synsets appearing in the
  glosses) have been shown to be always helpful. 
\item Graph algorithm. UKB implements different graph-based algorithms
  and variants to perform WSD. These are the main ones:
\\  \emph{ppr\_w2w}: apply personalized PageRank for each target
    word, that is, perform a random walk in the graph personalized on
    the word context. It yields the best results overall, at the cost
    of being more time consuming that the rest.
  \\ \emph{ppr}: same as above, but apply personalized PageRank to
    each sentence only once, disambiguating all content words in the
    sentence in one go. It is thus faster that the
    previous approach, but obtains worse results.
  \\ \emph{dfs}: unlike the two previous algorithms, which consider
    the WordNet graph as a whole, this algorithm first creates a
    subgraph for each context, following the method first presented
    in~\citet{NavigliLapata:10}, and then runs the PageRank algorithm
    over the subgraph. This option represents a compromise between
    \emph{ppr\_w2w} and \emph{ppr}, as it faster than than the
    former while better than the latter.
\item The PageRank algorithm has two parameters which were set as
  follows: number of iterations of power method (\emph{prank\_iter})
  30, and damping factor (\emph{prank\_damping}) 0.85.

\item Use of sense frequencies (\emph{dict\_weight}). Sense
  frequencies are a valuable piece of information that describe the
  frequencies of the associations between a word and its possible
  senses. The frequencies are often derived from manually sense
  annotated corpora, such as Semcor~\citep{semcor}. We use the sense
  frequency accompanying Wordnet, which, according to the
  documentation, ''represents the decimal number of times the sense is
  tagged in various semantic concordance texts''. The frequencies are
  smoothed adding one to all counts (\emph{dict\_weight\_smooth}). The
  sense frequency is used when initializing context words, and is also
  used to produce the final sense weights as a linear combination of
  sense frequencies and  graph-based sense
  probabilities. The use of sense frequencies with UKB was introduced in \cite{DBLP:journals/coling/AgirreLS14}.
\end{itemize}

%\todo[inline]{
%bestalde ondo egongo zen kontatzea, edo gutxienez nik jakitea, ukb bertsio bakoitzean zein ziren defektuzko parametroak, readme-etan ea zerbait edaten genuen eta 2009 eta 2014tan zer edaten genuen.}

\begin{table*}[t]
  \centering
%  \small
\begin{tabular}{lr|rrrrr}
                           & All           & S2    & S3    & S07  & S13  & S15  \\
  \hline
  \multicolumn{7}{c}{Single context sentence}                                                                            \\
  \hline
  ppr\_w2w                 & 66.9          & \textbf{69.0} & 65.7          & 53.9          & 67.1          & 69.9          \\
  dfs\_ppr                 & 65.2          & 67.5          & 65.6          & 53.6          & 62.7          & 68.2          \\
  ppr                      & 65.5          & 67.5          & \textbf{66.5} & \textbf{54.7} & 63.3          & 67.4          \\ \hline
  ppr\_w2w$_{\mathtt{nf}}$ & 60.2          & 63.7          & 55.1          & 42.2          & 63.5          & 63.8          \\
  ppr$_{\mathtt{nf}}$      & 57.1          & 60.5          & 53.8          & 41.3          & 58.0          & 61.4          \\
  dfs$_{\mathtt{nf}}$      & 58.7          & 63.3          & 52.8          & 40.4          & 61.6          & 62.5          \\
%  static                   & 57.7          & 62.9          & 51.9          & 41.3          & 59.9          & 60.1          \\
  \hline
  \multicolumn{7}{c}{One or more context sentences ($\# words \ge 20$)}                                                                            \\
  \hline
  ppr\_w2w                 & \textbf{67.3} & 68.8          & 66.1          & 53.0          & \textbf{68.8} & \textbf{70.3} \\
  ppr                      & 65.6          & 67.5          & 66.4          & 54.1          & 64.0          & 67.8          \\
  dfs                      & 65.7          & 67.9          & 65.9          & 54.5          & 64.2          & 68.1          \\ \hline
  ppr\_w2w$_{\mathtt{nf}}$ & 60.4          & 64.2          & 54.8          & 40.0          & 64.5          & 64.5          \\
  ppr$_{\mathtt{nf}}$      & 58.6          & 61.3          & 54.9          & 42.2          & 60.9          & 62.9          \\
  dfs$_{\mathtt{nf}}$      & 59.1          & 62.7          & 54.4          & 39.3          & 62.8          & 62.2          \\
%  static                   & 57.7          & 62.9          & 51.9          & 41.3          & 59.9          & 60.1          \\
%  dfs\_static             & 58.4          & 62.4          & 52.3          & 42.6          & 62.5          & 61.1          \\
  \hline
%  \multicolumn{7}{c}{Baseline}                                                                                             \\
%  \hline
%  UKB wn30g (ppr)         & 57.5          & 60.6          & 54.1          & 42.0          & 59.0          & 61.2          \\
%  MFS                      & 65.2          & 66.8          & 66.2          & \textbf{55.2} & 63.0          & 67.8          \\
\end{tabular}
\caption{Additional results on other settings of UKB. \textrm{nf} subscript stands for ``no sense frequency''. Top rows use a single sentence as context, while the bottom rows correspond to extended context (cf.  Sect. \ref{sec:ukb-param-sett}). Best results in bold.}
  \label{tab:main-results-ukb}
\end{table*}

\section{Comparison to the state-of-the-art}
\label{sec:comp-with-state}

We evaluate UKB on the recent evaluation dataset described
in~\cite{E17-1010}. This dataset comprises five standard English
all-words datasets, standardized into a unified format with gold keys
in WordNet version 3.0 (some of the original datasets used older
versions of WordNet). The dataset contains $7,253$ instances of
$2,659$ different content words (nouns, verbs, adjectives and
adverbs). The average ambiguity of the words in the dataset is of
$5.9$ senses per word. We report F1, the harmonic mean between
precision and recall, as computed by the evaluation code accompanying
the dataset.

%In our experiments, we use a graph comprising WordNet 3.0 relations,
%including gloss relations, and we pre-process input contexts as described
%above.

The two top rows in Table \ref{tab:comparison} show conflicting
results for UKB. The first row corresponds to UKB ran with the
settings described above. The second row was first reported in
\cite{E17-1010}. As the results show, that paper reports a suboptimal
use of UKB. In more recent work, \citet{aaai18} take up that result
and report it in their paper as well. The difference is of nearly 10
absolute F1 points overall.\footnote{Note that the UKB results for S2, S3 and
S07 (62.6, 63.0 and 48.6 respectively) are different from those in
\cite{DBLP:journals/coling/AgirreLS14}, which is to be expected, as
the new datasets have been converted to WordNet 3.0 (we confirmed
experimentally that this is the sole difference between the two experiments).} This decrease could be caused by the fact that  \citet{E17-1010} did not use sense frequencies. 

In addition to UKB, the table also reports the best performing
knowledge-based systems on this dataset. \citet{E17-1010} run several
well-known algorithms when presenting their datasets. We also report
\cite{aaai18}, the latest work on this area, as well as the most
frequent sense as given by WordNet counts (see Section
\ref{sec:ukb-param-sett}). The table shows that UKB yields the best
overall result. Note that \citet{Banerjee:2003:EGO:1630659.1630775} do not use sense frequency information.

For completeness, Table \ref{tab:comparison2} reports the results of supervised systems on
the same dataset, taken from the two works that use the dataset 
~\citep{C16-1130,raganato-dellibovi-navigli:2017:EMNLP2017}. As
expected, supervised systems outperform knowledge-based systems, by a small margin in some of the cases.

\section{Additional results}
\label{sec:results}

In addition to the results of UKB using the setting in
\cite{Agirre:09a,DBLP:journals/coling/AgirreLS14} as specified in
Section \ref{sec:ukb-param-sett}, we checked whether some reasonable
settings would obtain better results. Table \ref{tab:main-results-ukb}
shows the results when applying the three algorithms described in
Section \ref{sec:ukb-param-sett}, both with and without sense
frequencies, as well as using a single sentence for context or
extended context. The table shows that the key factor is the use of
sense frequencies, and systems that do not use them (those with a
\textrm{nf} subscript) suffer a loss between $7$ and $8$ percentage
points in F1. This would explain part of the decrease in performance
reported in \cite{E17-1010}, as they explicitly mention that they did
not activate the use of sense frequencies in UKB.

The table also shows that extending the context is mildly
effective. Regarding the algorithm, the table confirms that the best
method is \emph{ppr\_w2w}, followed by the subgraph approach
(\emph{dfs}) and \emph{ppr}.

\section{Conclusions}

This paper presents a case where an open-source NLP software was used
with suboptimal parameters by third parties. UKB was released with
suboptimal default parameters, and although the accompanying papers
did describe the necessary settings for good results on WSD, bad
results were not prevented. The results using the settings described
in the paper on newly released datasets show that UKB is the best
among knowledge-based WSD algorithms.

The take-away message for open-source NLP software authors is that
they should not rely on other researchers reading the papers with
care, and that it is extremely important to include, with the software
release, precise instructions and optimal default parameters, or
better still, end-to-end scripts that download all resources, perform
any necessary pre-processing and reproduce the results. UKB now
includes in version 3.1 such end-to-end scripts and the
appropriate default parameters.

\bibliographystyle{acl_natbib.bst}

\bibliography{main}

\begin{thebibliography}{22}
\expandafter\ifx\csname natexlab\endcsname\relax\def\natexlab#1{#1}\fi

\bibitem[{Agirre and Edmonds(2007)}]{Agirre:2007:WSD:1564561}
E.~Agirre and P.~Edmonds. 2007.
\newblock \emph{Word Sense Disambiguation: Algorithms and Applications}, 1st
  edition.
\newblock Springer Publishing Company, Incorporated.

\bibitem[{Agirre et~al.(2014)Agirre, de~Lacalle, and
  Soroa}]{DBLP:journals/coling/AgirreLS14}
E.~Agirre, O.~Lopez de~Lacalle, and A.~Soroa. 2014.
\newblock Random walks for knowledge-based word sense disambiguation.
\newblock \emph{Computational Linguistics}, 40(1):57--88.

\bibitem[{Agirre and Soroa(2009)}]{Agirre:09a}
E.~Agirre and A.~Soroa. 2009.
\newblock {Personalizing PageRank for Word Sense Disambiguation}.
\newblock In \emph{Proceedings of 14th Conference of the European Chapter of
  the Association for Computational Linguistics}, Athens, Greece.

\bibitem[{Agirre et~al.(2009)Agirre, Soroa, Alfonseca, Hall, Kravalova, and
  Pasca}]{agirre2009naacl}
E.~Agirre, A.~Soroa, E.~Alfonseca, K.~Hall, J.~Kravalova, and M.~Pasca. 2009.
\newblock {A Study on Similarity and Relatedness Using Distributional and
  {WordNet}-based Approaches}.
\newblock In \emph{Proceedings of annual meeting of the North American Chapter
  of the Association of Computational Linguistics (NAAC)}, Boulder, USA.

\bibitem[{Agirre et~al.(2010)Agirre, Soroa, and Stevenson}]{ASS2010}
E.~Agirre, A.~Soroa, and M.~Stevenson. 2010.
\newblock Graph-based word sense disambiguation of biomedical documents.
\newblock \emph{Bioinformatics}, 26:2889--2896.

\bibitem[{Agirre et~al.(2015)Agirre, Barrena, and Soroa}]{arxiv}
Eneko Agirre, Ander Barrena, and Aitor Soroa. 2015.
\newblock \href {http://arxiv.org/abs/1503.01655} {Studying the wikipedia
  hyperlink graph for relatedness and disambiguation}.
\newblock In \emph{ArXiv repository}.

\bibitem[{Banerjee and Pedersen(2003)}]{Banerjee:2003:EGO:1630659.1630775}
Satanjeev Banerjee and Ted Pedersen. 2003.
\newblock \href {http://dl.acm.org/citation.cfm?id=1630659.1630775} {Extended
  gloss overlaps as a measure of semantic relatedness}.
\newblock In \emph{Proceedings of the 18th International Joint Conference on
  Artificial Intelligence}, IJCAI'03, pages 805--810, San Francisco, CA, USA.
  Morgan Kaufmann Publishers Inc.

\bibitem[{Basile et~al.(2014)Basile, Caputo, and
  Semeraro}]{conf/coling/BasileCS14}
Pierpaolo Basile, Annalina Caputo, and Giovanni Semeraro. 2014.
\newblock \href
  {http://dblp.uni-trier.de/db/conf/coling/coling2014.html#BasileCS14} {An
  enhanced lesk word sense disambiguation algorithm through a distributional
  semantic model.}
\newblock In \emph{COLING}, pages 1591--1600. ACL.

\bibitem[{Bond and Paik(2012)}]{bond2012survey}
Francis Bond and Kyonghee Paik. 2012.
\newblock A survey of wordnets and their licenses.
\newblock In \emph{GWC 2012 6th International Global Wordnet Conference}.

\bibitem[{Chaplot and Sakajhutdinov(2018)}]{aaai18}
D.S. Chaplot and R.~Sakajhutdinov. 2018.
\newblock {Knowledge-based Word Sense Disambiguation using Topic Models}.
\newblock In \emph{AAAI}.

\bibitem[{Erbs et~al.(2012)Erbs, Agirre, Soroa, Barrena, Etxebarria, Gurevych,
  and Zesch}]{tac12}
Nicolai Erbs, Eneko Agirre, Aitor Soroa, Ander Barrena, Ugaitz Etxebarria,
  Iryna Gurevych, and Torsten Zesch. 2012.
\newblock Ukp-ubc entity linking at tac-kbp.
\newblock In \emph{Text Analysis Conference, Knowledge Base Population}.

\bibitem[{Fellbaum(1998)}]{fellbaum98wordnet}
Christiane Fellbaum, editor. 1998.
\newblock \emph{{WordNet: an electronic lexical database}}.
\newblock MIT Press.

\bibitem[{Goikoetxea et~al.(2015)Goikoetxea, Agirre, and
  Soroa}]{goikoetxea2015}
Josu Goikoetxea, Eneko Agirre, and Aitor Soroa. 2015.
\newblock Random walks and neural network language models on knowledge bases.
\newblock In \emph{Proceedings of the Annual Meeting of the North American
  chapter of the Association of Computational Linguistics (NAACL HLT 2015),
  pages 1434-1439. ISBN: 978-1-937284-73-2. Denver (USA).}

\bibitem[{Iacobacci et~al.(2016)Iacobacci, Pilehvar, and
  Navigli}]{iacobacci-pilehvar-navigli:2016:P16-1}
Ignacio Iacobacci, Mohammad~Taher Pilehvar, and Roberto Navigli. 2016.
\newblock Embeddings for word sense disambiguation: An evaluation study.
\newblock In \emph{Proceedings of the 54th Annual Meeting of the Association
  for Computational Linguistics (Volume 1: Long Papers)}, pages 897--907,
  Berlin, Germany. Association for Computational Linguistics.

\bibitem[{Melamud et~al.(2016)Melamud, Goldberger, and
  Dagan}]{Melamud2016context2vecLG}
Oren Melamud, Jacob Goldberger, and Ido Dagan. 2016.
\newblock context2vec: Learning generic context embedding with bidirectional
  lstm.
\newblock In \emph{CoNLL}.

\bibitem[{Miller et~al.(1993)Miller, Leacock, Tengi, and Bunker}]{semcor}
George~A. Miller, Claudia Leacock, Randee Tengi, and Ross~T. Bunker. 1993.
\newblock \href {https://doi.org/http://dx.doi.org/10.3115/1075671.1075742} {A
  semantic concordance}.
\newblock In \emph{Proceedings of the workshop on Human Language Technology},
  HLT '93, pages 303--308, Stroudsburg, PA, USA. Association for Computational
  Linguistics.

\bibitem[{Moro et~al.(2014)Moro, Raganato, and Navigli}]{Moro:2014:ELmeetsWSD}
A.~Moro, A.~Raganato, and R.~Navigli. 2014.
\newblock Entity linking meets word sense disambiguation: a unified approach.
\newblock \emph{Transactions of the Association of Computational Linguistics},
  2:231--244.

\bibitem[{Navigli and Lapata(2010)}]{NavigliLapata:10}
R.~Navigli and M.~Lapata. 2010.
\newblock {An Experimental Study of Graph Connectivity for Unsupervised Word
  Sense Disambiguation}.
\newblock \emph{IEEE Transactions on Pattern Analysis and Machine
  Intelligence}, 32(4):678--692.

\bibitem[{Raganato et~al.(2017{\natexlab{a}})Raganato, Camacho-Collados, and
  Navigli}]{E17-1010}
Alessandro Raganato, Jose Camacho-Collados, and Roberto Navigli.
  2017{\natexlab{a}}.
\newblock \href {http://www.aclweb.org/anthology/E17-1010} {{Word Sense
  Disambiguation: A Unified Evaluation Framework and Empirical Comparison}}.
\newblock In \emph{Proceedings of the 15th Conference of the European Chapter
  of the Association for Computational Linguistics: Volume 1, Long Papers},
  pages 99--110. Association for Computational Linguistics.

\bibitem[{Raganato et~al.(2017{\natexlab{b}})Raganato, Delli~Bovi, and
  Navigli}]{raganato-dellibovi-navigli:2017:EMNLP2017}
Alessandro Raganato, Claudio Delli~Bovi, and Roberto Navigli.
  2017{\natexlab{b}}.
\newblock Neural sequence learning models for word sense disambiguation.
\newblock In \emph{Proceedings of the 2017 Conference on Empirical Methods in
  Natural Language Processing}, pages 1156--1167, Copenhagen, Denmark.
  Association for Computational Linguistics.

\bibitem[{Yuan et~al.(2016)Yuan, Richardson, Doherty, Evans, and
  Altendorf}]{C16-1130}
Dayu Yuan, Julian Richardson, Ryan Doherty, Colin Evans, and Eric Altendorf.
  2016.
\newblock \href {http://www.aclweb.org/anthology/C16-1130} {Semi-supervised
  word sense disambiguation with neural models}.
\newblock In \emph{Proceedings of COLING 2016, the 26th International
  Conference on Computational Linguistics: Technical Papers}, pages 1374--1385.
  The COLING 2016 Organizing Committee.

\bibitem[{Zhong and Ng(2010)}]{Zhong:2010}
Zhi Zhong and Hwee~Tou Ng. 2010.
\newblock \href {http://dl.acm.org/citation.cfm?id=1858933.1858947} {It makes
  sense: A wide-coverage word sense disambiguation system for free text}.
\newblock In \emph{Proceedings of the ACL 2010 System Demonstrations}, ACLDemos
  '10, pages 78--83, Stroudsburg, PA, USA. Association for Computational
  Linguistics.

\end{thebibliography}

\end{document}